\documentclass[10pt,twocolumn,letterpaper]{article}

\usepackage{iccv}
\usepackage{times}
\usepackage{epsfig}
\usepackage{graphicx}
\usepackage{amsmath}
\usepackage{amssymb}
\usepackage{subfigure}
\usepackage[normalem]{ulem}
\useunder{\uline}{\ul}{}

\usepackage[pagebackref=true,breaklinks=true,letterpaper=true,colorlinks,bookmarks=false]{hyperref}

\iccvfinalcopy 

\def\iccvPaperID{3346} 

\ificcvfinal\fi

\begin{document}

\title{Prompt Tuning based Adapter for Vision-Language Model Adaption}

\author{Jingchen Sun\\
University at Buffalo\\
{\tt\small jsun39@buffalo.edu}
\and
Jiayu Qin\\
University at Buffalo\\
{\tt\small jiayuqin@buffalo.edu}
\and
Zihao Lin\\
Duke University\\
{\tt\small zihao.lin@duke.edu}
\and
Changyou Chen\\
University at Buffalo\\
{\tt\small cchangyou@gmail.com}
}

\maketitle
\ificcvfinal\thispagestyle{empty}\fi

\begin{abstract}
Large pre-trained vision-language (VL) models have shown significant promise in adapting to various downstream tasks. However, fine-tuning the entire network is challenging due to the massive number of model parameters. To address this issue, efficient adaptation methods such as prompt tuning have been proposed. We explore the idea of prompt tuning with multi-task pre-trained initialization and find it can significantly improve model performance. Based on our findings, we introduce a new model, termed Prompt-Adapter, that combines pre-trained prompt tunning with an efficient adaptation network. Our approach beat the state-of-the-art methods in few-shot image classification on the public 11 datasets, especially in settings with limited data instances such as 1 shot, 2 shots, 4 shots, and 8 shots images. Our proposed method demonstrates the promise of combining prompt tuning and parameter-efficient networks for efficient vision-language model adaptation. 
The code is publicly available at: \href {https://github.com/Jingchensun/prompt_adapter}{https://github.com/Jingchensun/prompt\_adapter}
\end{abstract}

\section{Introduction}

In recent years, there has been a growing interest in developing Vision-Language (VL) models that can perform joint reasoning over visual and textual information. Large-scale VL models, such as CLIP \cite{clip} and ALIGN \cite{jia2021scaling}, have shown impressive zero-shot transfer learning ability on downstream tasks \cite{downstream-seg, downstream-video}, including image classification \cite{krizhevsky2017imagenet, coop, elevater} and open-vocabulary object detection \cite{ od-gu2021open, od-zang2022open, od-zhou2022detecting, elevater}. These models are pre-trained on web-scale images and text pairs \cite{clip} and contain a lot of cross-domain knowledge. However, adapting these large-scale VL models to downstream tasks is still quite challenging due to the size and complexity of the models.

Several methods have been proposed to adapt large pre-trained VL models. One of the most commonly used methods is fine-tuning, which involves updating the model's parameters on the task-specific dataset. However, fine-tuning the entire network is computationally expensive and may lead to overfitting, especially when the target dataset is small. As an alternative, two methods have been proposed: prompt tuning \cite{prompt1, prompt2, prompt3} and parameter-efficient \cite{parameter1, parameter2} tuning. Prompt tuning involves adding an extra set of words or learnable parameters that are fed into the text encoder, allowing the model to obtain task-specific outputs. In contrast, parameter-efficient tuning involves adding an extra network or parameters to learn the representation of the downstream tasks, which reduces the computational cost of model adaptation.

Prompt learning is a technique used to fine-tune pre-trained language models for specific downstream tasks. It involves providing a prompt, which is a natural language statement or question that constrains the model to generate a specific output. Prompt learning has shown great promise in improving the zero-shot transfer learning ability of large-scale VL models. There are mainly three types of prompt tuning methods: text prompt tuning, visual prompt tuning, and unified prompt tuning. The representative work of text prompt tuning is CoOp \cite{coop}. CoOp injects additional text as input into the text encoder to help guide the model toward the desired output. Visual prompt tuning works, like VPT \cite{vpt} and visual prompting \cite{visual-prompt}, by injecting additional parameters into multiple layers of the vision transformer to optimize the image features output. Unified prompt tuning \cite{upt} combines text and visual prompts for a better trade-off between the two. However, prompt-based methods are highly impacted by the dataset and need longer training time to achieve optimal results.

Tip-Adapter \cite{tip-adpter} is a recent method proposed for adapting the CLIP model to new downstream tasks in a training-free manner. The approach appends a non-parametric cache model to the weight-frozen CLIP model, where the cache model stores few-shot visual features encoded by CLIP and their ground-truth labels under one-hot encodings. During inference, the cache model is used to retrieve the few-shot knowledge and incorporate it with CLIP's pre-trained knowledge to achieve high performance on downstream tasks. Tip-Adapter has the advantage of not requiring fine-tuning or additional training for adapting CLIP to new tasks, which significantly reduces computational costs. However, it has the drawback of relying on manual prompts for the text encoder, which may not fully capture the knowledge of the CLIP text encoder.

In this research, we present a novel approach that integrates prompt tuning and parameter-efficient networks to overcome the limitations of each individual method. Specifically, we propose to generalize the manual prompting of the Tip-Adapter with learnable prompts adapted from CoOp's text prompt. We demonstrate the effectiveness of our approach in few-shot image classification. Our methods have two variants called Prompt-Adapter and Prompt-Adapter-F. Prompt-Adapter is a variant that is only based on the prior knowledge of pre-trained text prompt and cache model, it does not need training and can outperform the previous. Furthermore, we extend Prompt-Adapter by utilizing a multi-task trained prompt to initialize the text prompt and achieve a further 1.55\% improvement in classification accuracy. For another variant of our work Prompt-Adapter-F, we train the network with 20 epochs and obtain a 0.49\% improvement in accuracy compared to previous methods, setting a new state-of-the-art on few-shot classification. Our results suggest that the integration of prompt tuning and parameter-efficient networks can enhance the efficiency and performance of VL model adaptation for image classification.

In summary, we propose a novel approach that combines multi-task pre-trained prompt learning with parameter-efficient networks to achieve efficient few-shot image classification. Our contributions include 1) showing the effectiveness of the multi-task pre-trained prompt mechanism in improving single-task prompt recognition accuracy, 2) proposing a new network architecture that incorporates prompt learning and a parameter-efficient network, and 3) demonstrating the superiority of our approach through few-shot image classification experiments on 11 datasets. The results show that our approach outperforms state-of-the-art models in terms of accuracy, especially in extreme situations where data are limited available. Our method highlights the potential of combining prompt learning and parameter-efficient networks for efficient vision-language model adaptation.

\section{Related Work}
\textbf{Vision Language Model.}
Vision and language models have become increasingly popular in recent years and have shown remarkable success in various computer vision tasks \cite{coop, od-gu2021open, od-zang2022open, od-zhou2022detecting, downstream-seg, downstream-video}. These models typically consist of an image encoder and a text encoder, trained using contrastive loss on large-scale image-text pairs. CLIP \cite{clip} and ALIGN \cite{jia2021scaling} are some of the most prominent models in this domain. CLIP (Contrastive Language-Image Pre-training) is a large-scale vision language model trained on 400 million image-text pairs that have demonstrated impressive transferability in cross-domain downstream tasks. In this context, our work focuses on transferring CLIP into 11 cross-domain image classification tasks. The integration of prompt learning-based methods and parameter-efficient methods has also shown significant improvement in few-shot performance with small computing resources, making it an area of active research.

\textbf{Parameter Tuning}
Large pre-trained vision language models like CLIP have achieved state-of-the-art performance in various downstream tasks, but their high computational requirements make them difficult to deploy in resource-constrained environments. To address this issue, several approaches \cite{parameter1, clip-adapter, tip-adpter} have been proposed to reduce the number of parameters and computations required for inference while maintaining high accuracy. CLIP-Adapter \cite{clip-adapter} appends a lightweight two-layer Multi-Layer Perceptron (MLP) to the pre-trained weight-fixed CLIP model and optimizes its parameters via stochastic gradient descent (SGD). Tip Adapter \cite{tip-adpter} constructs a non-parametric cache model that stores features of training images and their labels as a key-value database. By aggregating the information from the cache model with the text features, Tip-Adapter significantly boosts the classification accuracy without training over Zero-shot CLIP. However, Tip-Adapter still uses the manual Prompt as an image classifier, which can not fully utilize the huge knowledge of the text encoder of the CLIP model.

\textbf{Prompt Tuning}
Prompt engineering \cite{prompt1, prompt2, prompt3} has been widely used in natural language processing (NLP) to improve the performance of language models on specific tasks. A prompt is a piece of text that is added to the input to guide the model toward a particular output. Prompts can be used to provide additional information to the model, such as a task description or a set of constraints. Prompt engineering involves designing effective prompts that can guide the model toward the desired output. For example, by feeding a manual text prompt "a photo of a \{\}" to the text encoder \cite{clip}, the CLIP model has shown strong zero-shot image recognization ability.

Meanwhile, prompt tuning aims to learn an optimal prompt automatically through fine-tuning or meta-learning. In recent studies, several methods have been proposed to improve the efficiency and effectiveness of prompt engineering and learning. For example, CoOp \cite{coop} introduced a collaborative optimization method that jointly optimizes prompts and model parameters. On the other hand, Visual Prompt Tuning (VPT) \cite{vpt} uses a visual prompt that is tailored to the image input, improving the model's performance on image classification tasks. Unified Prompt Tuning (UPT) \cite{upt} is a recent method that learns a unified prompt for multiple tasks, which leads to better performance in downstream tasks. These methods have shown promising results in improving the performance of vision and language models, and we aim to build upon their success in our work.

\textbf{Few-shot Learning}
Few-shot learning has been a challenging research topic in image classification, where the objective is to recognize novel classes from only a few training examples \cite{dhillon2019baseline, tian2020rethinking, afrasiyabi2022matching, liu2022few}. This problem has garnered significant attention in the computer vision community, as it is more realistic and practical in many real-world scenarios where labeled data is scarce or costly to obtain. In recent years, various methods have been proposed for few-shot image classification.

One of the earliest approaches for few-shot learning was Siamese neural networks \cite{koch2015siamese}, which used a distance metric to compare images and learn to distinguish between classes. Later, the idea of meta-learning was introduced \cite{ren2018meta}, which trains a model to learn how to learn from few examples. This approach led to the development of popular few-shot learning algorithms like Matching Networks \cite{cai2018memory}, Prototypical Networks \cite{snell2017prototypical, deng2020meta}, and Relation Networks \cite{sung2018learning}. These methods have shown promising results on various benchmark datasets, but they usually rely on simple feature extractors and do not scale well to large-scale image classification problems.

Recently, CLIP's \cite{clip} ability to jointly reason about text and images has inspired a new line of research on few-shot learning for vision and language models. Various adaptation methods, such as CoOp \cite{coop}, UPT \cite{upt}, and Tip-Adapter \cite{tip-adpter} have been proposed to fine-tune the CLIP model on few-shot datasets. These methods have achieved state-of-the-art results on several benchmark datasets, demonstrating the effectiveness of adapting large-scale vision and language models to few-shot learning tasks.

\section{Method}
We first revisit the CLIP, CoOp, and Tip-Adapter in Section~\ref{sec:pre}, then present the technical details of our proposed method in Section~\ref{sec:method}.

\subsection{Preliminaries}\label{sec:pre}
\textbf{CLIP} \cite{clip} model is a large-scale neural network model, which is trained on a diverse set of image-text pairs to learn the relationship between the visual and textual features. Unlike traditional vision models, CLIP leverages the massive amount of textual and visual information available on the internet to learn more robust and accurate representations. 

The CLIP model consists of two encoders: an image encoder and a text encoder, which are trained together using contrastive loss to project images and the corresponding text descriptions into a common embedding space. CLIP jointly trains an image encoder $\psi $ and a text encoder $\phi$ and uses a symmetric contrastive loss to match the batch of image-text pairs. The training objective $L_{ \rm CLIP}$ is:

\begin{equation}
 L_{\text CLIP} = L_{i2t} + L_{t2i}
\end{equation}
with $L_{i2t}$ representing an image-to-text contrastive loss and $L_{t2i}$ a text-to-image contrastive loss. The contrastive loss is calculated as follows:
\begin{equation}
L_{i2t} = -\sum_{i\epsilon \ss }^{}\log\frac{\exp \ (\cos \ (u_{i}, v_{i})/\tau )}{\sum _{j\epsilon \ss }\exp \ (\cos \ (u_{i}, v_{j})/\tau )}
\end{equation}

\begin{equation}
L_{t2i} = -\sum_{j\epsilon \ss }^{}\log\frac{\exp \ (\cos \ (u_{j}, v_{j})/\tau )}{\sum _{j\epsilon \ss }\exp \ (\cos \ (u_{i}, v_{j})/\tau )}
\end{equation}
where $u = \psi (x)$ represents the projection of image $x$ to the final hidden space, $v=\phi (y)$ indicates the projection of text y to the final embedding. $cos$ denotes the cosine similarity; $\tau $ is a learnable temperature value.

 The joint contrastive learning allows the CLIP to learn to associate textual descriptions with visual features, enabling it to perform a variety of tasks such as image classification \cite{krizhevsky2017imagenet}, object detection \cite{od-zhou2022detecting, od-gu2021open}, segmentation \cite{downstream-seg}, etc.

\textbf{CoOp}
Context Optimization \cite{coop} is a simple approach that is specifically designed for adapting large-scale vision-language models for downstream image recognition tasks. CoOp is built on the principle of optimizing the context, or prompt, to improve the model's ability to recognize images. The method is based on the insight that the context or prompt can significantly impact the performance of the model. CoOp aims to find the optimal prompt by iteratively refining it based on the performance of the model on the downstream task.

CoOp utilized a number of unified contexts to model context words with continuous vectors for downstream image recognition. It is the first prompt-based network used in the CLIP model adapting. They designed the unified context vectors in the text encoder of the CLIP model and shares the same context with all classes. The prompt given to the text encoder $ g\left ( \cdot  \right ) $ is designed with the following form,

\begin{equation}
    t=[\text V]_1[\text V]_2...[\text V]_M[\text Class]
\end{equation}

where each $[\text V]_1[\text V]_2...[\text V]_M$ is a vector with the same dimension as word embeddings (i.e., 512 for CLIP), and $M$ is a hyperparameter determining the number of context tokens.
The CoOp approach has demonstrated impressive results in adapting CLIP-like vision-language models for downstream image recognition through its use of text prompt learning vectors. Despite this, one of the major drawbacks of CoOp is its relatively long training time, typically requiring 200 epochs to achieve optimal performance compared to other methods. Therefore, there is a need for further research to explore ways to reduce the training time of CoOp while maintaining its effectiveness in adapting large vision-language models.

\begin{figure*}[htb]
  \centering
  \includegraphics[width=0.9\linewidth]{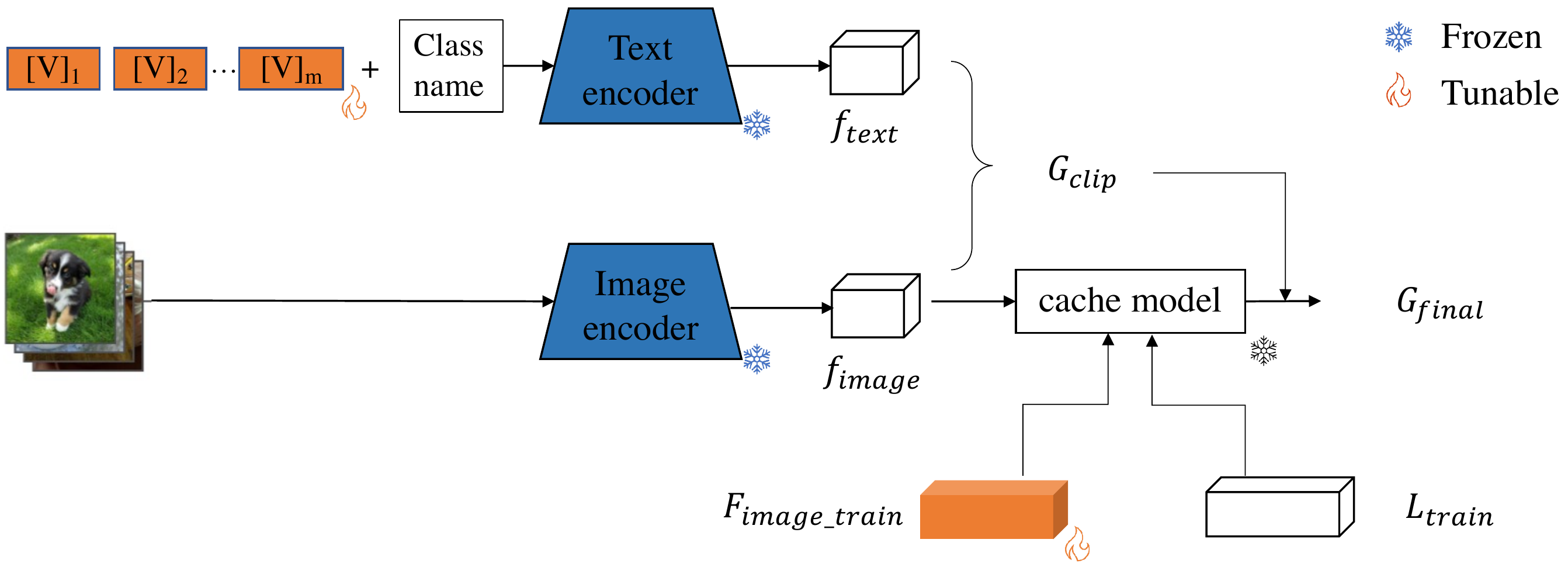}
  \caption{An illustration of our method Prompt Adapter. The learned prompt and the class name are sent to the frozen CLIP text encoder. While the test images are sent to the frozen CLIP image encoder. The $G_\text{clip}$ is obtained by calculating the cosine similarity between text embeddings and image features. Further, the $G_\text{cache}$ is obtained by the cache model. The Final logit $G_\text{final}$ is a linear combination of $G_\text{cache}$ and $G_\text{clip}$.  The $G_\text{final}$ is used to classify the images.}
  \label{fig:intro}
\end{figure*}

\textbf{Tip-Adapter} \cite{tip-adpter} is a recent parameter-efficient approach for adapting CLIP-like vision-language models to downstream image recognition tasks. Unlike traditional fine-tuning methods, Tip-Adapter constructs a non-parametric cache model that stores the features of training images and their corresponding labels as a key-value database. Given $K$-shot images with $N$-class training samples, we use $I_{K}$ to represent the image features of $K$-shot images and $L_{N}$ to represent their labels. The key in Tip-Adapter is the pre-trained $N*K$ image feature $\text {F}_{\text{train}}$ extracted by the CLIP image encoder, while the value is an $N$-dimensional one-hot vector $\text {L}_{\text{train}}$, which represents the ground truth label. The $\text {F}_{\text{train}}$ and $\text {L}_{\text{train}}$ are represented by:

\begin{equation}
\text {F}_{\text{train}}=\text{VisualEncoder}(I_{K})
\end{equation}
\begin{equation}
\text {L}_{\text{train}} = \text{OneHot}(L_{N})
\end{equation}

After constructing the cache model, the adaption of CLIP can be simply achieved by two matrix-vector multiplications $f_{\text{test}} {\text F}_{\text{train}}$. The term $f_{\text{test}} {\text F}_{\text{train}}^{T}$ is equivalent to the cosine similarities between test feature $f_{\text{test}}$ and all few-shot training features ${\text F}_{\text{train}}$. Then, the prediction for the cache model can be obtained via a linear combination of cached values weighted $ A \text {L}_{\text{train}}$ and the original CLIP logits. The $A$ is calculated as:
\begin{equation}
A = \exp \ (-\beta (1-f_{\text{test}} {\text F}_{\text{train}}^{T}))
\end{equation}

and the final output logits of the test image by Tip-Adapter are then calculated as:

\begin{equation}
{\text {Logits}} = \alpha (A) {\text L_{\text{train}}} + f_{\text{test}}W_{c}^{T}
\end{equation}

 By aggregating the information from the cache model with the text features, Tip-Adapter can significantly boost classification accuracy without the need for further training on the CLIP model. This approach leverages the prior knowledge encoded in CLIP by feature retrieval and can incorporate new knowledge from the few-shot training set. 
 
When given more shots, Tip-Adapter lags behind the training-required methods without training. Thus the author proposed an augmented version called Tip-Adapter-F \cite{tip-adpter}, which treats the keys in the cache model as learnable parameters, and fine-tunes them via SGD. Tip-Adapter-F achieves state-of-the-art performance on the few-shot adapting of the CLIP model. However, Tip-Adapter still relies on manual prompts as image classifiers, limiting its ability to fully utilize the vast knowledge of the text encoder of the CLIP model.

\subsection{Our Method}\label{sec:method} 
Tip-Adapter represents the state-of-the-art few-shot learning approach for CLIP-based vision-language models. However, it still relies on manual prompts for image classification and is limited in its ability to fully utilize the vast knowledge of the CLIP text encoder. Our proposed Prompt-Adapter overcomes this limitation by incorporating text prompt learning with Tip-Adapter, which can achieve better few-shot learning performance. As shown in Figure 1, our Prompt-Adapter adopts a similar network structure as Tip-Adapter but replaces the original manual prompt "a photo of a {}" with learnable vectors $[\text V]_1[\text V]_2...[\text V]_M$. By doing so, our approach can optimize the text prompt to better capture the essential features of the few-shot training set and improve classification accuracy.

Our Prompt-Adapter has two variants. One is the training-free variant, denoted as Prompt-Adapter; and the other is the learnable variant, denoted as Prompt-Adapter-F. 

In \textbf{Prompt-Adapter}, we do not need the training phase, and only rely on the cache model and fixed prompt. We first construct the cache model with the few-shot train images and the one-hot ground truth label. And then we directly use the learned prompt from CoOp. 

When testing the model, the learned prompt and class names are sent to the text encoder to obtain the text embeddings $f_{\text{text}}$. And in the image branch, the test images are sent to the CLIP image encoder and obtained image features $f_{\text{image\_test}}$. Then we calculate the clip logits $G_\text{clip}$, which is the cosine similarity between $f_{\text{text}}$ and $f_{\text{image\_test}}$, given by:
\begin{equation}
G_\text{clip} = f_{\text{text}}\cdot f_{\text{image\_test}}
\end{equation}
The $\text{clip\ logits}$ $ G_\text{clip}$ can be used to select the most matched image and label pairs for classification. In practice, we do not only rely on the clip logits to classify images. Thus, we further calculate the cache logits $G_\text{cache}$, which is the cosine similarity between the test image features and the cached image features. The cache logits is given by:
\begin{equation}
\begin{split}
G_\text {cache} = \exp \ (-\beta (1-f_{\text{image\_test}}\cdot \text {F}_{\text{image\_train}}^{T})) \text{L}_{\text{train}}
\end{split}
\end{equation}
where $\text {F}_{\text{image\_train}}$ and $\text{L}_{\text{train}}$ represent the train images and the one-hot labels in the cache model. And $f_{\text{image\_test}}\cdot \text {F}_{\text{image\_train}}$ represent the cosine similarity between test image features and trained image features; an exponential function is applied to convert similarities into positive values; and $\beta$ stands for a modulating hyperparameter to control the degree of sharpness \cite{tip-adpter}. The final logits are a linear combination of cache logits and clip logits, given by:
\begin{equation}
{G_\text {final}} = \alpha \cdot {G_\text {cache}} + {G_\text {clip}}
\label{eqution-L}
\end{equation}
where $\alpha$ is the weight coefficient to balance the information from the training data and the prior knowledge of the CLIP model \cite{tip-adpter}. The final Logits can be used to select the most matched image and text pairs. Thus we can use the final Logits to realize image classification. 

In \textbf{Prompt-Adapter-F }, the image features in the cache model are learnable. There are training phases and test phases in the method. In the training phase, training images are sent to the image encoder, and the output of the image encoder is denoted as $f_{\text{image\_train}}$. The cache logits is adapted to:
\begin{equation}
{G_\text {cache}} = \exp \ (-\beta (1-f_{\text{image\_train}} \cdot \text {F}_{\theta }^{T})) \text {L}_{\text{train}}
\end{equation}
where $F_{\theta }$ represents the learned parameters of cache features. The final logits of the Prompt Adapter F are the same as equation \ref{eqution-L}. Thus, the learned parameters can be optimized by the cross-entropy loss, here $G_\text{Target}$ refers to ground truth label logits: 
\begin{equation}
L_{\text {prompt\_adapter}} = CE( G_\text {final}, G_\text{Target})
\end{equation}
where $CE$ represents the cross entropy loss function.
When training Prompt-Adapter-F, two learnable parameters are present in the network: the text prompt $[\text V]_1[\text V]_2...[\text V]_M$ and the cache features $F_{\theta }$. To optimize these parameters, we employ two different strategies: the separately optimized strategy and the joint optimization strategy. In the separately optimized strategy, we first optimize the text prompt. Specifically, the learnable vectors $[\text V]_1[\text V]_2...[\text V]_M$ are optimized by the cross entropy loss function between clip logits and ground truth labels. The loss function $L_{\text {prompt}}$ is given by:
\begin{equation}
L_{\text {prompt}} = CE (G_\text{clip}, G_\text{Target})
\end{equation}

After we trained the prompt, we froze the parameters of the text prompt, and then optimize $\text {F}_{\theta }$ until we obtain the best performance. On the other hand, the joint optimization strategy directly optimizes the final output logits $G_\text {final}$. Because $G_\text {final}$ is a linear combination with the clip logits $G_\text {clip}$ and the cache logits $G_\text {cache}$. When we use the cross-entropy function to optimize the final output logits, the optimizer will automatically update the parameter in the text prompt and the cache features at the same time. Results from our ablation study will show that the separately optimized strategy performs better than the joint optimization strategy. Hence, we use the separately optimized strategy as the default strategy in our experiment.

\paragraph{Multi-task Initialization}
An important phenomenon we found is that the prompt initialization method has a significant impact on the performance of the model. Our default initialization approach involves randomly initializing the text prompt and using few-shot images and labels to train the prompt. To improve model performance, inspired by previous works \cite{shen2022multitask}, we first use 11 datasets as the source tasks, we train the shareable prompt among the 11 datasets. And then we use the shareable prompt as initialization and adapt the network to a single task. For the single-task prompt learning, we directly optimize the prompt with the cross-entropy loss function of each task. Once the sign-task prompt is learned, the single-task prompt will be used as the final text prompt in our network.

\section{Experiment}
To comprehensively compare our methods with other works, we first conducted detailed experiments for 20-shot image classification, where there are only 20 labeled examples available for each class.  Then, we extended the experiment set to include results for 1, 2, 4, 8, and 16 shots. Finally, we performed an ablation study to investigate the effects of different initialization methods and training strategies on the performance of the network.
\subsection{Few Shot Image Classfication}
\textbf{Datasets.} Following previous works such as CoOp \cite{coop} and Tip-Adapter \cite{tip-adpter}, we use 11 publicly available image classification datasets. These datasets include ImageNet \cite{imagenet}, Caltech101 \cite{caltech101}, OxfordPets \cite{oxfordpets}, StanfordCars \cite{standfordcars}, Flowers102 \cite{flowers102}, Food101 \cite{food101}, FGVCAircraft \cite{fgv}, SUN397 \cite{sun397}, DTD \cite{dtd}, EuroSAT \cite{eurosat}, and UCF101 \cite{eurosat}. By selecting a diverse range of datasets, we aimed to ensure that the evaluation was comprehensive and our methods can generalize across different domains.

ImageNet \cite{imagenet} is a massive dataset containing over 14 million images with more than 20,000 object categories. Caltech101 \cite{caltech101} is a smaller dataset consisting of 101 object categories, containing about 9,000 images. The OxfordPets \cite{oxfordpets} dataset includes over 7,000 images of pets in various categories, such as cats, dogs, and birds. StanfordCars \cite{standfordcars} is a dataset of cars consisting of over 16,000 images of 196 car models. Flowers102 \cite{flowers102} is a dataset with over 8,000 flower images in 102 categories. Food101 \cite{food101} is a dataset consisting of 101 food categories, including pizza, sushi, and burgers. FGVCAircraft \cite{fgv} is a dataset containing over 10,000 images of aircraft, including commercial and military planes. SUN397 \cite{sun397} is a scene recognition dataset that includes over 130,000 images of indoor and outdoor scenes. DTD (Descriptive Textures) \cite{dtd} is a dataset with 47 texture categories, including fabrics, tiles, and plants. EuroSAT \cite{eurosat} is a dataset with 27,000 satellite images of ten land-use classes. Finally, UCF101 \cite{ucf101} is a dataset with 13,000 videos in 101 action categories, such as playing basketball or brushing teeth. Table 1 is the statistics of these 11 datasets.

\begin{table}[htb]
\centering
\caption{The statistics of these 11 image classification datasets.}
\begin{tabular}{@{}lllll@{}}
\hline
                       & classes & train  & val    & test   \\ \hline
Oxford\_Pets            & 37      & 2,944  & 736    & 3,669  \\
Flowers102         & 102     & 4,093  & 1,633  & 2,463  \\
Fgvc\_Aircraft          & 100     & 3,334  & 3,333  & 3,333  \\
Describable   Textures & 47      & 2,820  & 1,128  & 1,692  \\
Eurosat                & 10      & 13,500 & 5,400  & 8,100  \\
Stanford\_Cars          & 196     & 6,509  & 1,635  & 8,041  \\
Food101                & 101     & 50,500 & 20,200 & 30,300 \\
Sun397                 & 397     & 15,880 & 3,970  & 19,850 \\
Caltech101             & 100     & 4,128  & 1,649  & 2,465  \\
Ucf101                 & 101     & 7,639  & 1,898  & 3,783  \\
ImageNet               & 1,000   & 1.28M  & N/A    & 50,000 \\ \hline
\end{tabular}
\end{table}

\textbf{Training Details.}
When training the learned prompt $[\text V]_1[\text V]_2...[\text V]_M$, the maximum training epoch of the ImageNet dataset is fixed at 50. While for the other 10 datasets, the maximum training epoch is set to 200 epochs for 16/8 shots, 100 epochs for 4/2 shots, and 50 epochs for 1 shot. The optimizer is SGD and the learning rate is 0.002 with batch size 32. And the learning rate is scheduled by the cosine annealing rule. We follow all other settings as the default settings in CoOp \cite{coop} for fairly comparing.

While training for the cache features $F_{\theta }$, we follow the default settings on Tip-Adapter \cite{tip-adpter} and set 100-epoch training for the EuroSAT dataset and only 20-epoch training for the other 10 datasets. All the experiments are done on 1, 2, 4, 8, 16, and 20 shots training sets, and test on the full test sets. The initial learning rate is 0.001 with a batch size of 256, and the AdamW optimizer with a cosine scheduler.

\textbf{Baseline}. We compare our method with various existing methods to evaluate its effectiveness. In particular, we use the following five methods as baselines for comparison purposes: 1) Zero-shot CLIP \cite{clip}: This baseline method relies solely on the CLIP model without any parameter fine-tuning. The hand-crafted text prompt template "a photo of a \{\}" is used as the text encoder input. 2) Linear-probe CLIP \cite{clip}: This method involves adding an additional linear classifier on top of the feature extraction layer of the frozen CLIP model. The classifier is trained on a few-shot training set to learn the corresponding labels. 3) CoOp \cite{coop}: This method aims to optimize the learnable vectors as the text prompt. Specifically, it employs a context optimization technique to improve the performance of the CLIP model. 4) UPT \cite{upt}: This method presents a unified text and visual prompt tuning approach. It employs a lightweight self-attention network to generate the prompt for CLIP's text and visual encoders. 5) Tip-Adapter \cite{tip-adpter}: This method is based on a key-value cache model that stores and retrieves knowledge. It is a training-free adaptation method that enables parameter-efficient adaptation of large vision and language models.

For fair comparisons, we use the same backbone with ViT-B16, and data preprocessing with CLIP. We report the average results of three random seed-running experiments to reduce variance and increase result robustness. In addition, we adopt the default settings for each baseline method as described in the original papers.

\begin{table*}[htb]
\centering
\caption{Few-shots image classification for different methods.}
\label{table:main}
\resizebox{\textwidth}{16mm}{

\begin{tabular}{lcccccccccccc}
\hline
                      & Oxford\_pets & Flowers102 & FGVCAircraft & DTD   & EuroSAT & StanfordCars & Food101 & SUN397 & caltech101 & ucf101 & ImageNet & Average \\ \hline
Zero-Shot Clip        & 89.13      & 70.65         & 24.87        & 44.03 & 48.26   & 65.55        & 85.88   & 62.58  & 93.27      & 67.70  & 68.79    & 65.52   \\
CoCp                  & 91.53      & 96.40         & 40.30        & 69.47 & 84.00   & 79.20        & 85.00   & 74.43  & 95.70      & 82.40  & 71.60    & 79.09   \\
Tip-Adapter           & 91.88      & 94.60         & 39.96        & 66.08 & 78.01   & 75.39        & 86.45   & 71.94  & 95.09      & 78.51  & 70.32    & 77.11   \\
Prompt-Adapter        & 88.12      & 96.55         & 43.08        & 70.33 & 82.86   & 79.49        & 83.76   & 72.91  & 94.93      & 80.23  & 70.63    & \textbf{78.44}   \\ \hline
Tip-Adapter-F         & 93.08      & 96.31         & 45.45        & 71.93 & 87.43   & 83.82        & 87.38   & 76.30  & 95.94      & 85.01  & 73.45    & 81.46   \\
Unifed Prompt Tunning & 92.95      & 97.11         & 46.80        & 70.65 & 90.51   & 84.33        & 85.00   & 75.92  & 95.94      & 84.03  & 72.63    & 81.44   \\
Prompt-Adapter-F      & 92.40      & 98.05         & 49.08        & 71.45 & 87.56   & 83.39        & 86.96   & 75.81  & 95.66      & 83.95  & 72.36    & \textbf{81.52}   \\ \hline

\end{tabular}}
\end{table*}

\begin{figure*}[htbp]
\centering
\subfigure
{
    \begin{minipage}[b]{.3\linewidth}
        \centering
        \includegraphics[scale=0.34]{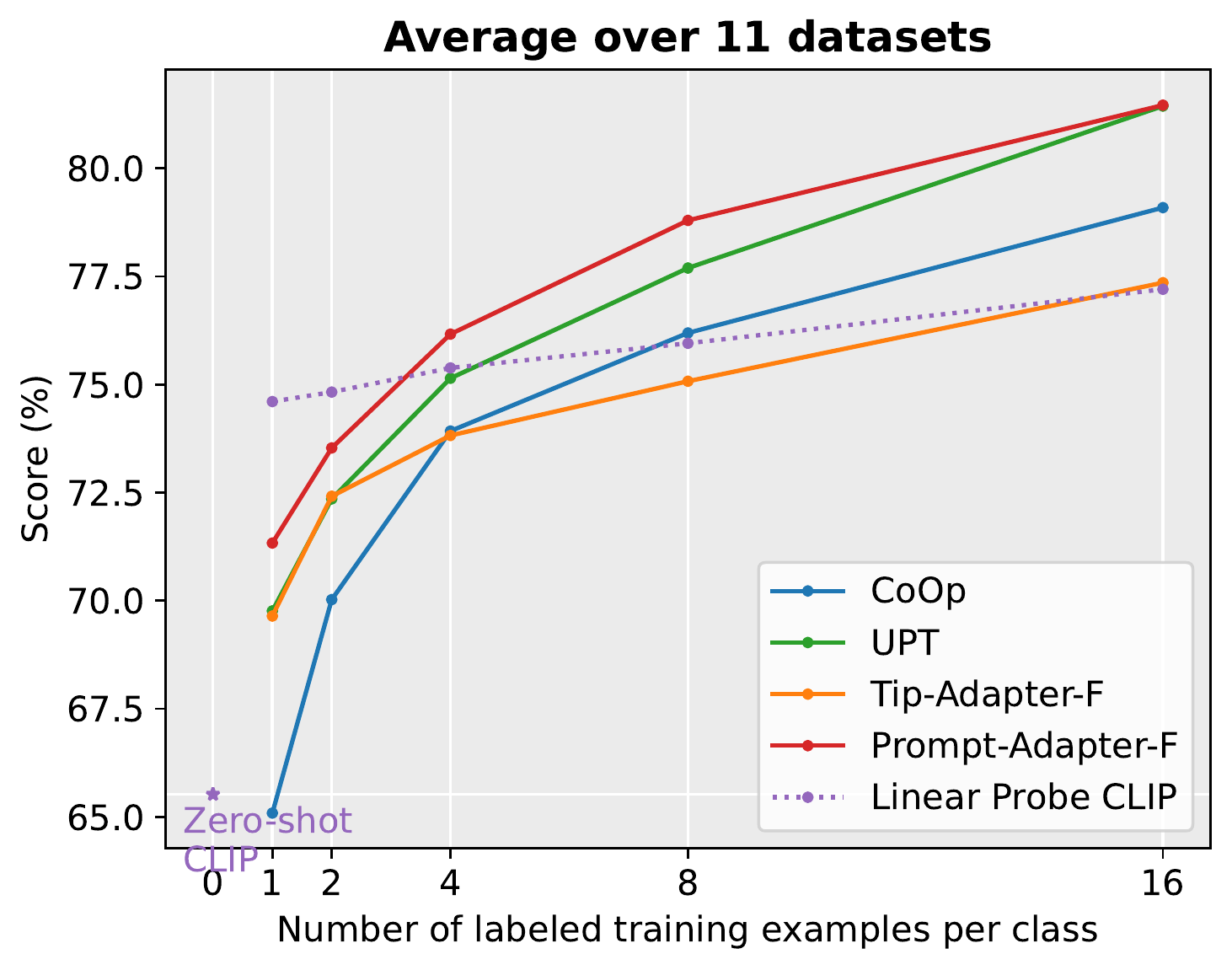}
        \label{fig:b}
    \end{minipage}
}
\subfigure
{
 	\begin{minipage}[b]{.3\linewidth}
        \centering
        \includegraphics[scale=0.34]{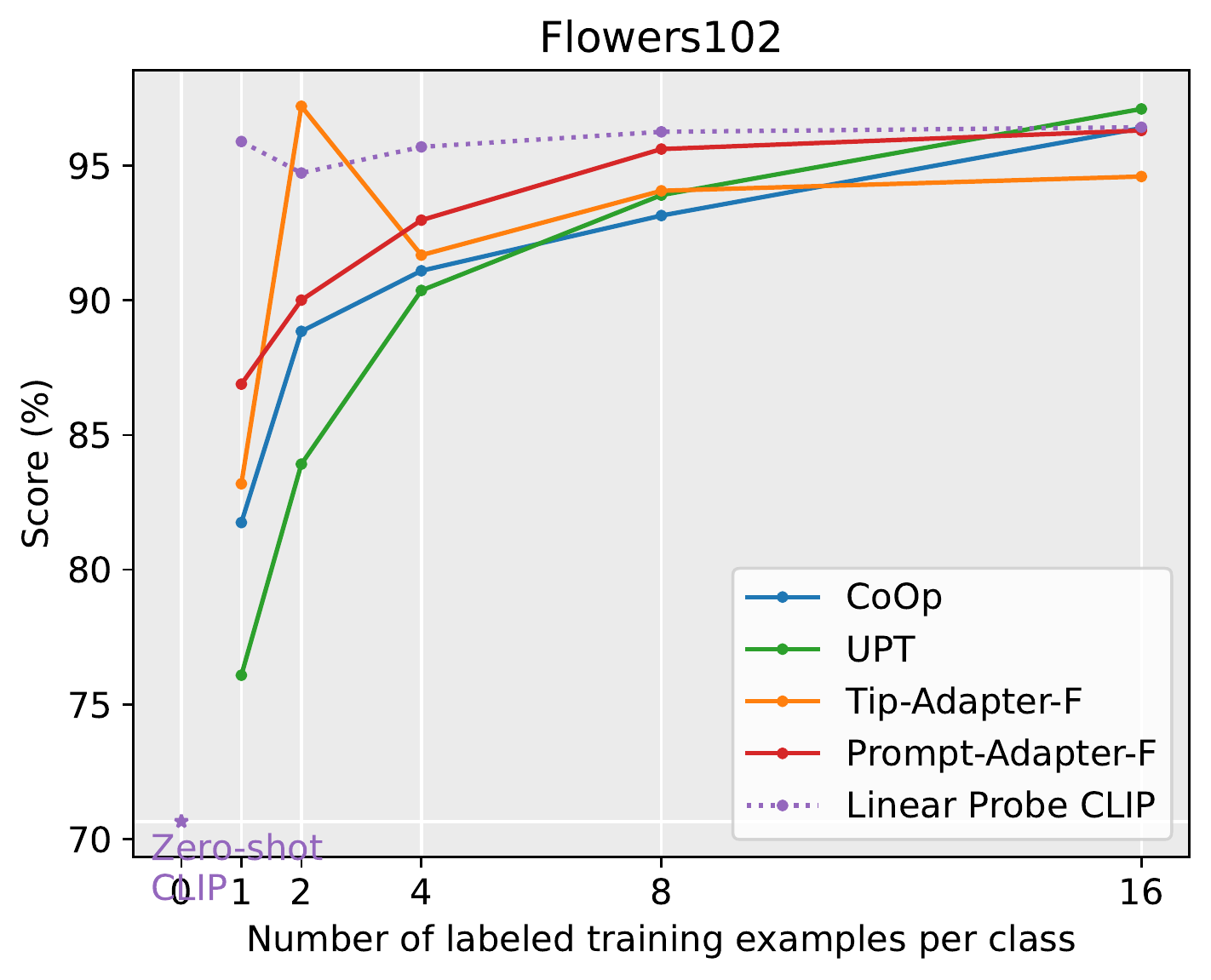}
        \label{fig:b}
    \end{minipage}
}
\subfigure
{
 	\begin{minipage}[b]{.3\linewidth}
        \centering
        \includegraphics[scale=0.34]{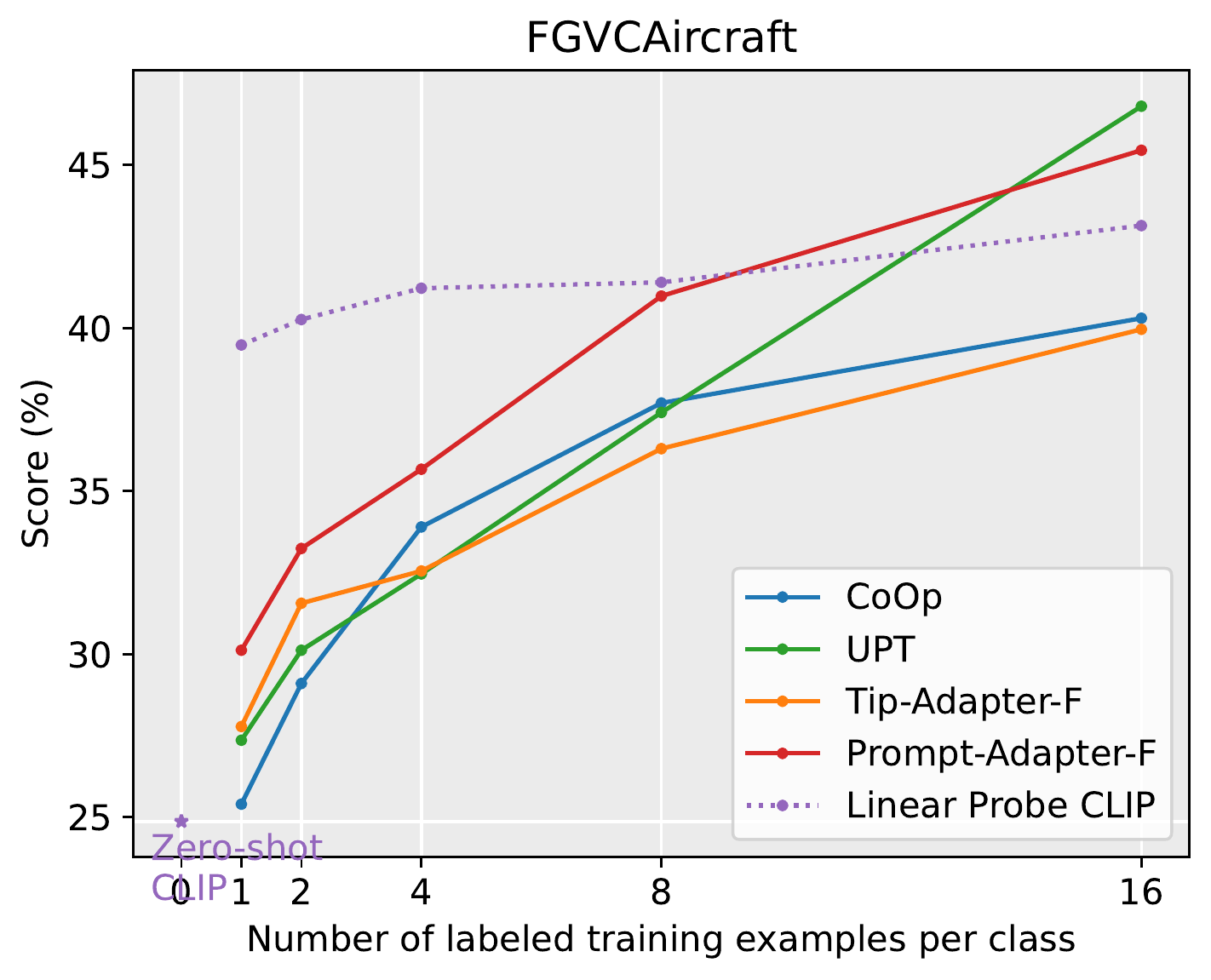}
        \label{fig:b}
    \end{minipage}
}
\subfigure
{
    \begin{minipage}[b]{.3\linewidth}
        \centering
        \includegraphics[scale=0.34]{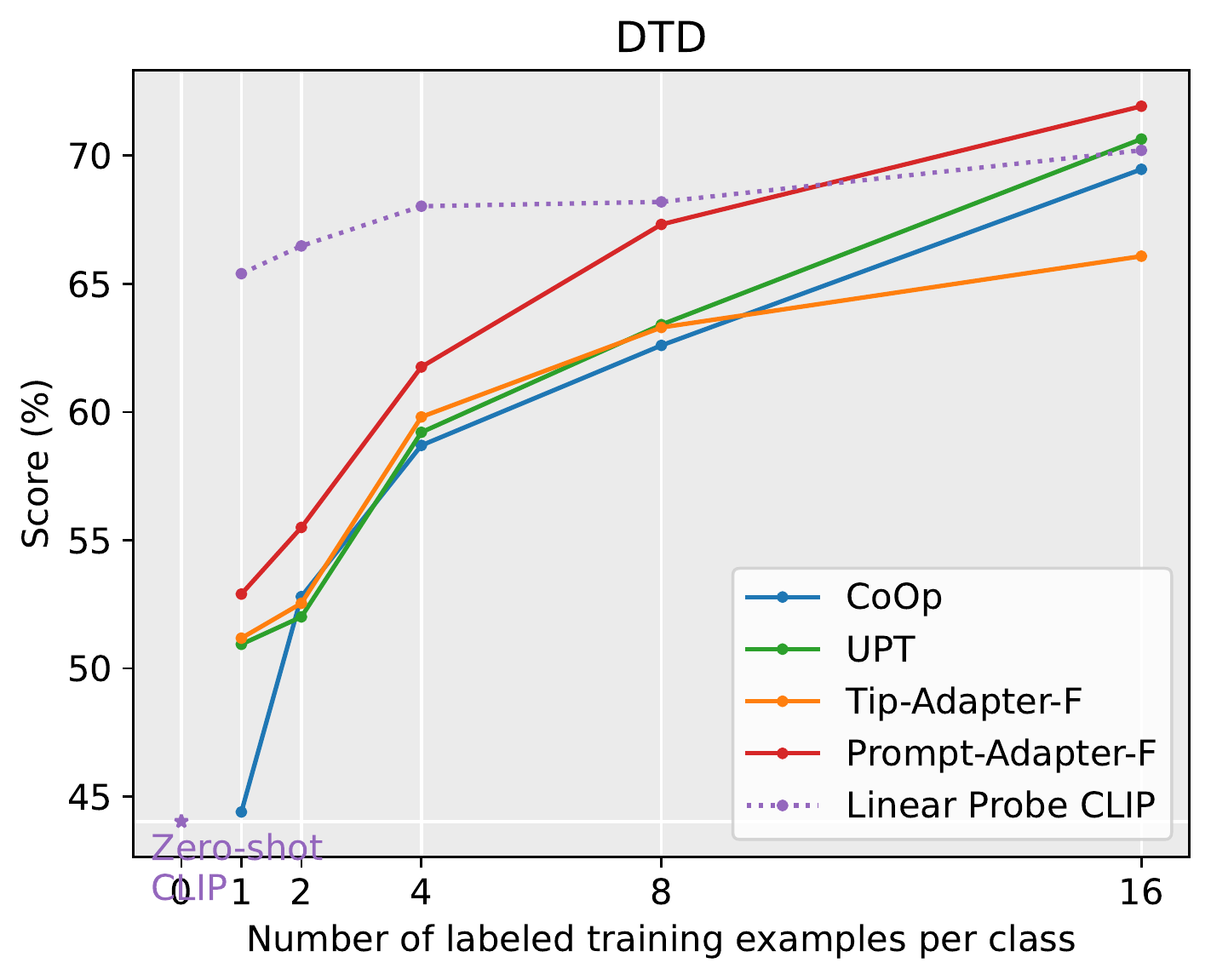}
        \label{fig:b}
    \end{minipage}
}
\subfigure
{
 	\begin{minipage}[b]{.3\linewidth}
        \centering
        \includegraphics[scale=0.34]{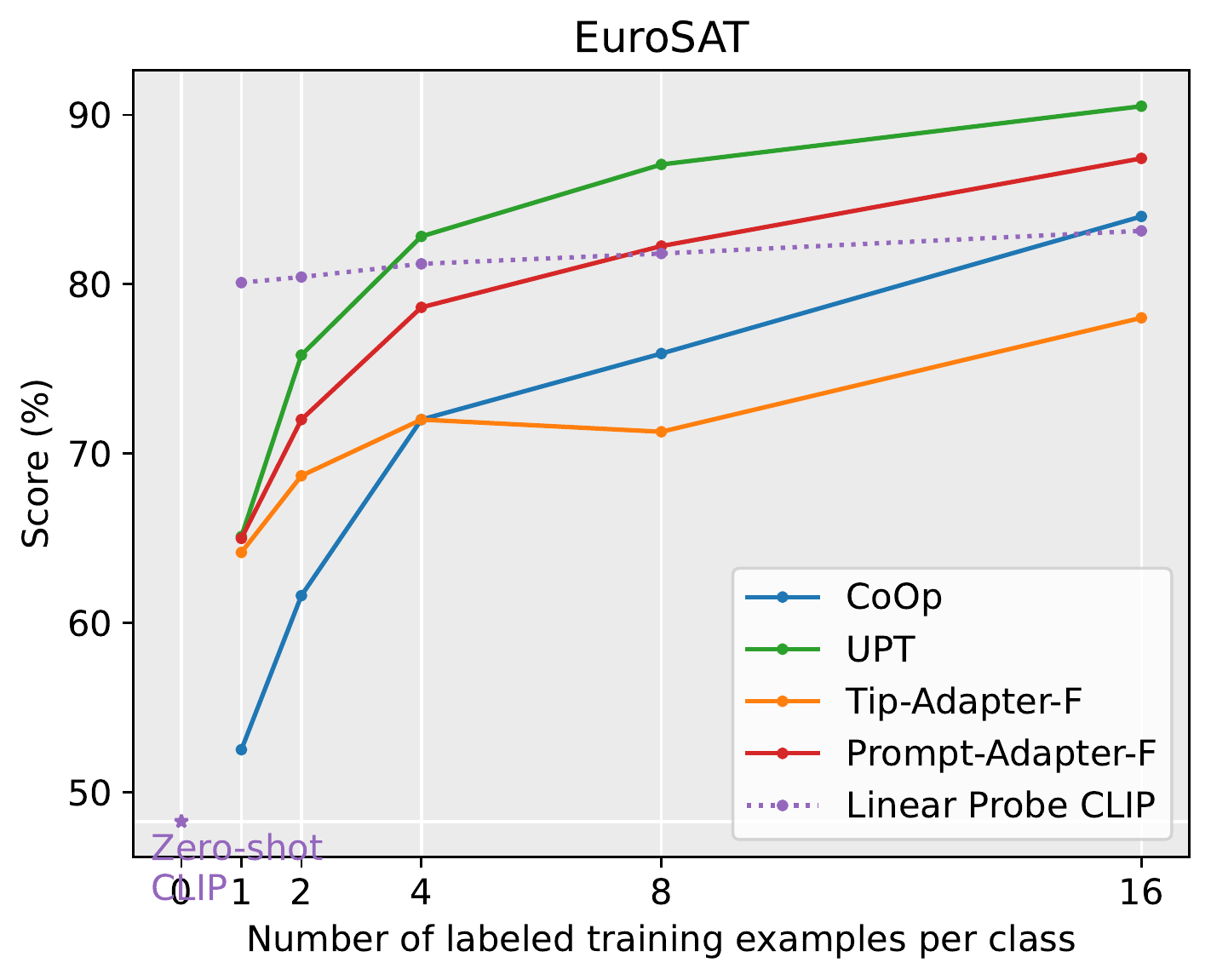}
        \label{fig:b}
    \end{minipage}
}
\subfigure
{
 	\begin{minipage}[b]{.3\linewidth}
        \centering
        \includegraphics[scale=0.34]{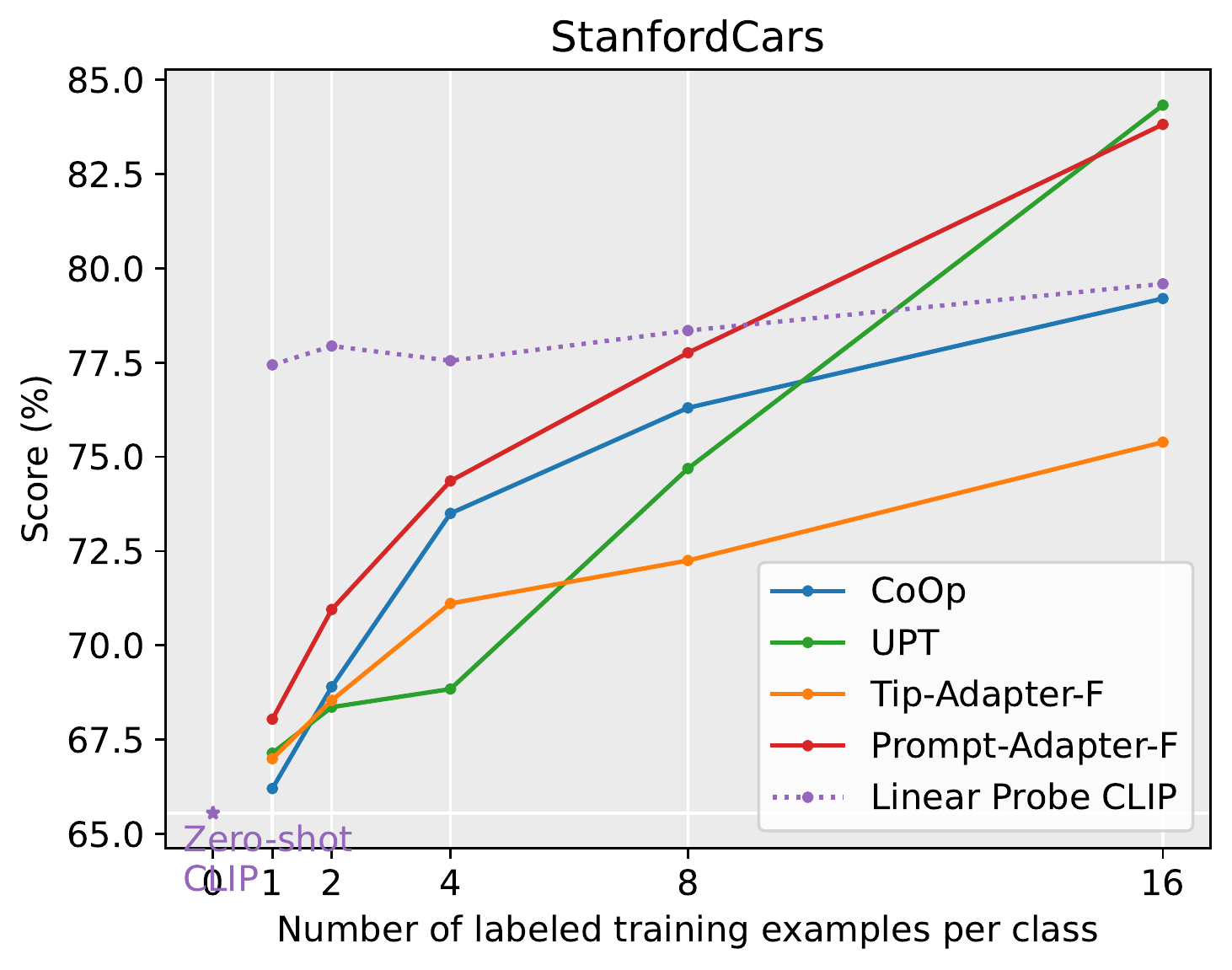}
        \label{fig:b}
    \end{minipage}
}
\subfigure
{
    \begin{minipage}[b]{.3\linewidth}
        \centering
        \includegraphics[scale=0.34]{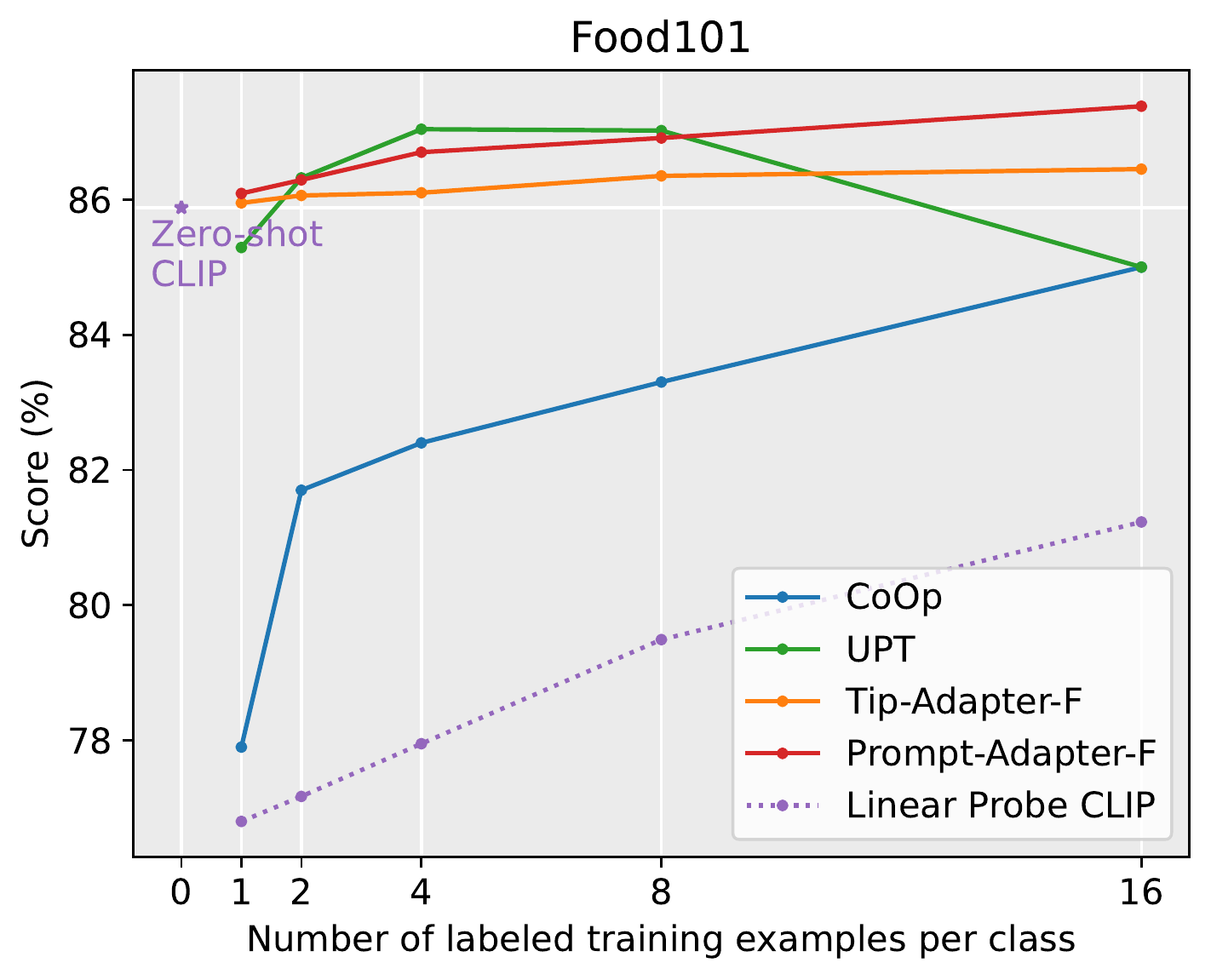}
        \label{fig:b}
    \end{minipage}
}
\subfigure
{
 	\begin{minipage}[b]{.3\linewidth}
        \centering
        \includegraphics[scale=0.34]{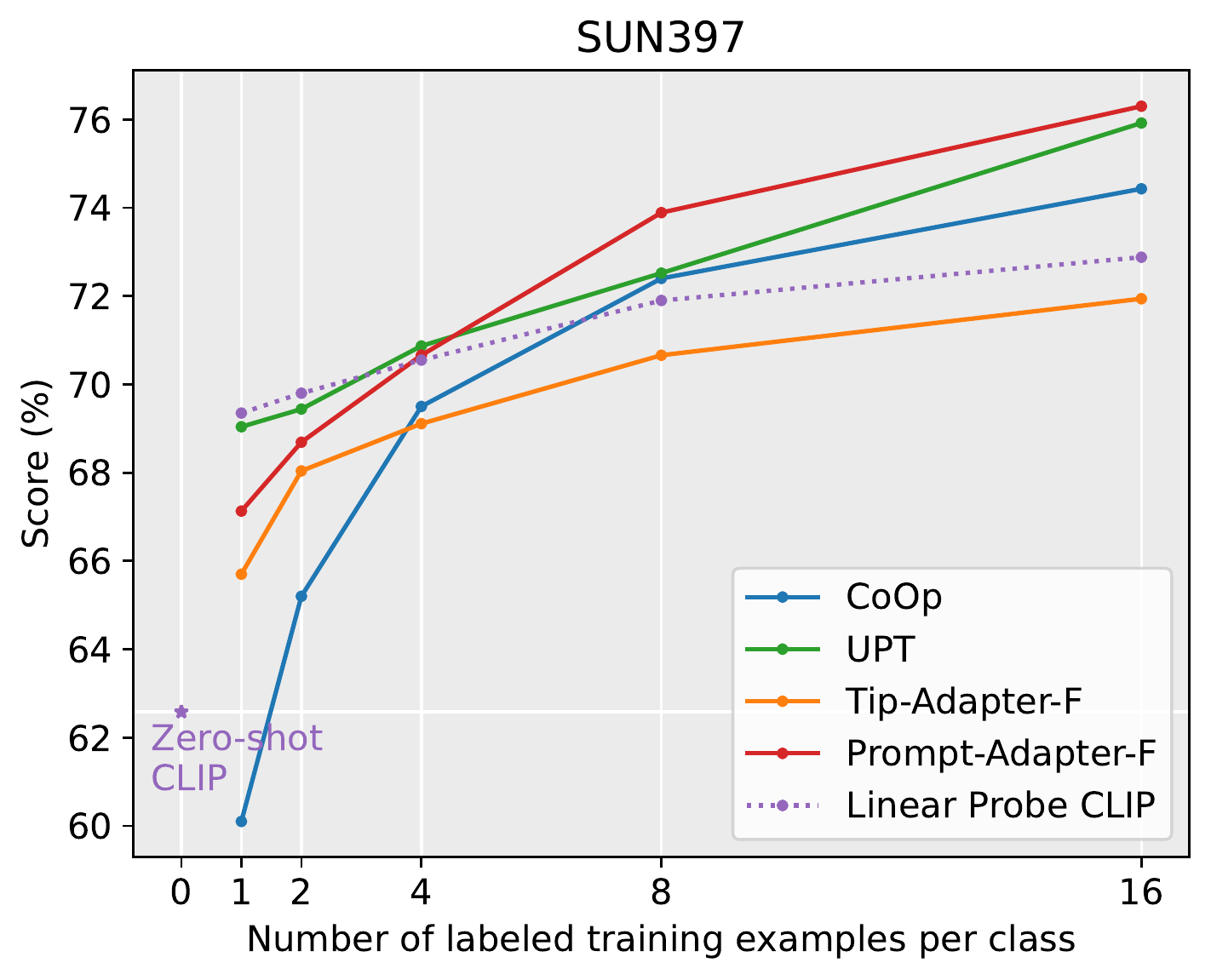}
        \label{fig:b}
    \end{minipage}
}
\subfigure
{
 	\begin{minipage}[b]{.3\linewidth}
        \centering
        \includegraphics[scale=0.34]{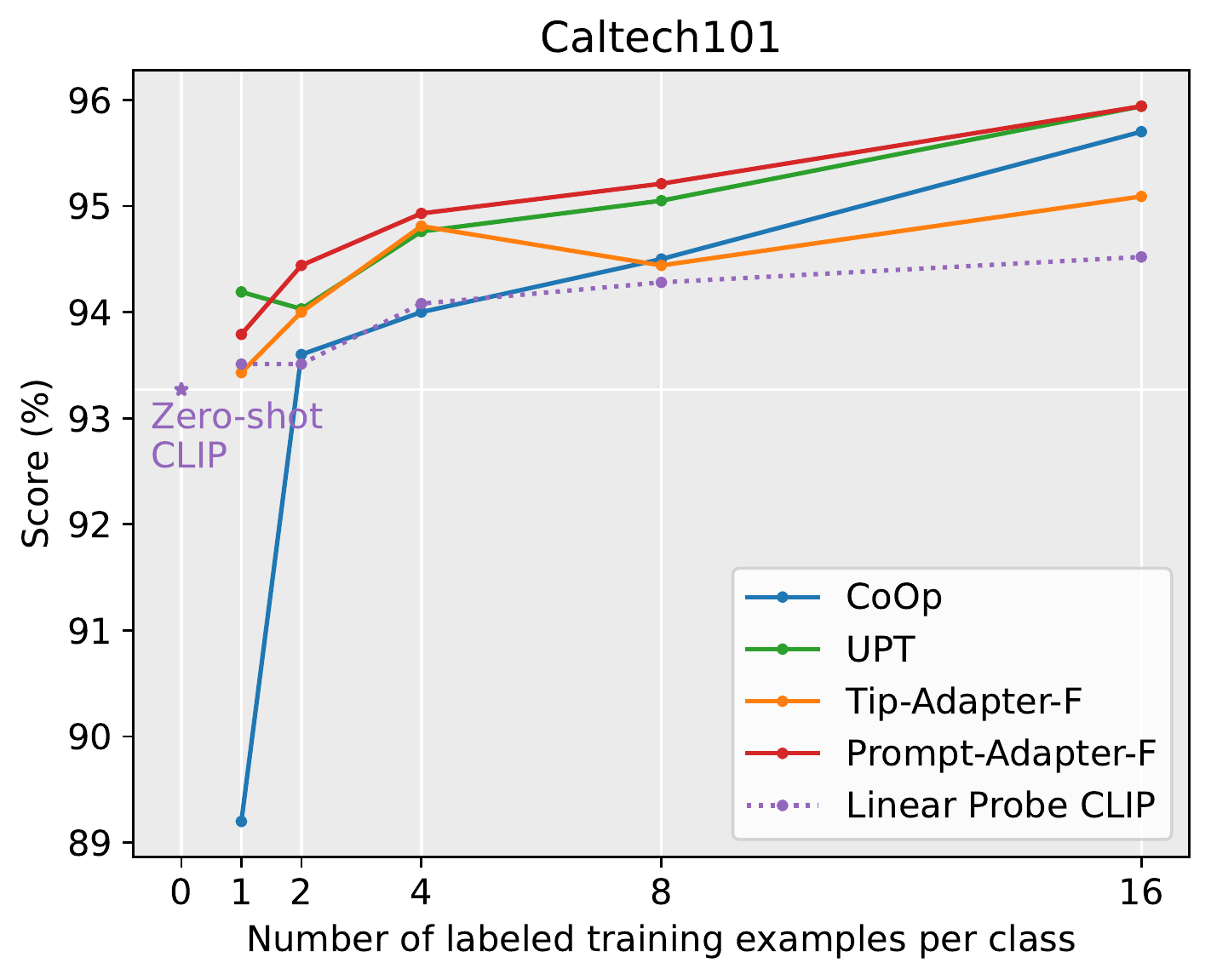}
        \label{fig:b}
    \end{minipage}
}
\subfigure
{
    \begin{minipage}[b]{.3\linewidth}
        \centering
        \includegraphics[scale=0.34]{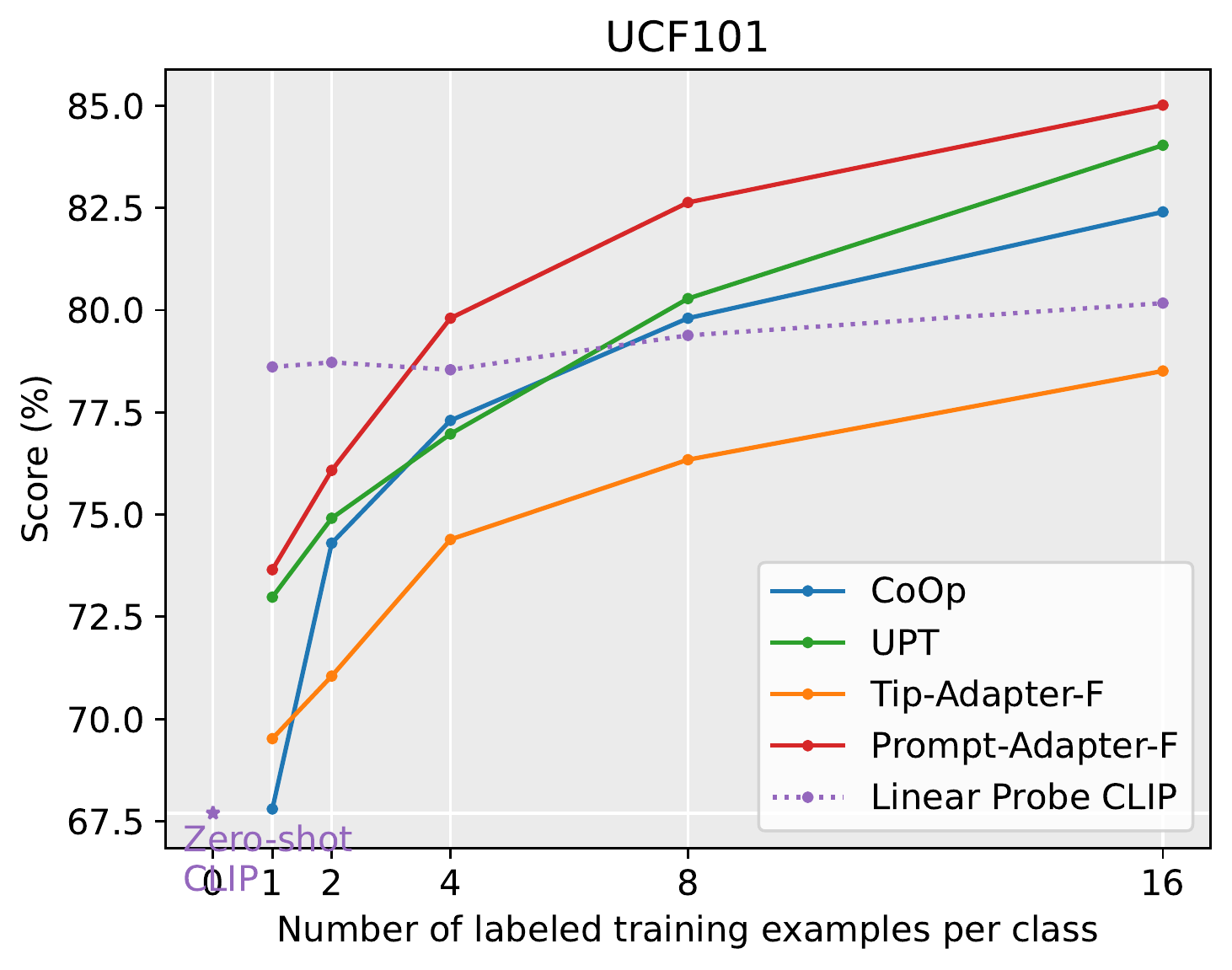}
        \label{fig:b}
    \end{minipage}
}
\subfigure
{
 	\begin{minipage}[b]{.3\linewidth}
        \centering
        \includegraphics[scale=0.34]{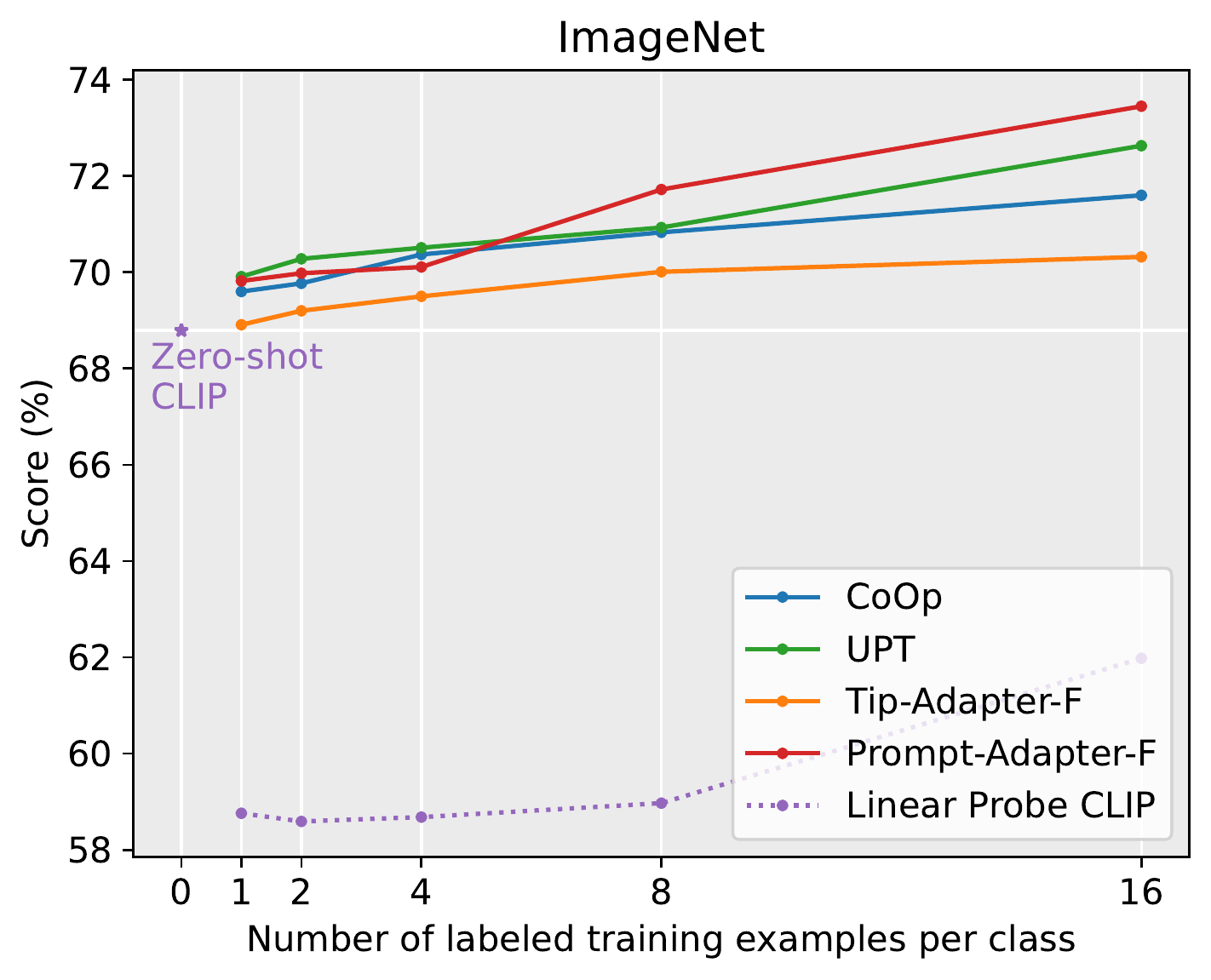}
        \label{fig:b}
    \end{minipage}
}
\subfigure
{
 	\begin{minipage}[b]{.3\linewidth}
        \centering
        \includegraphics[scale=0.34]{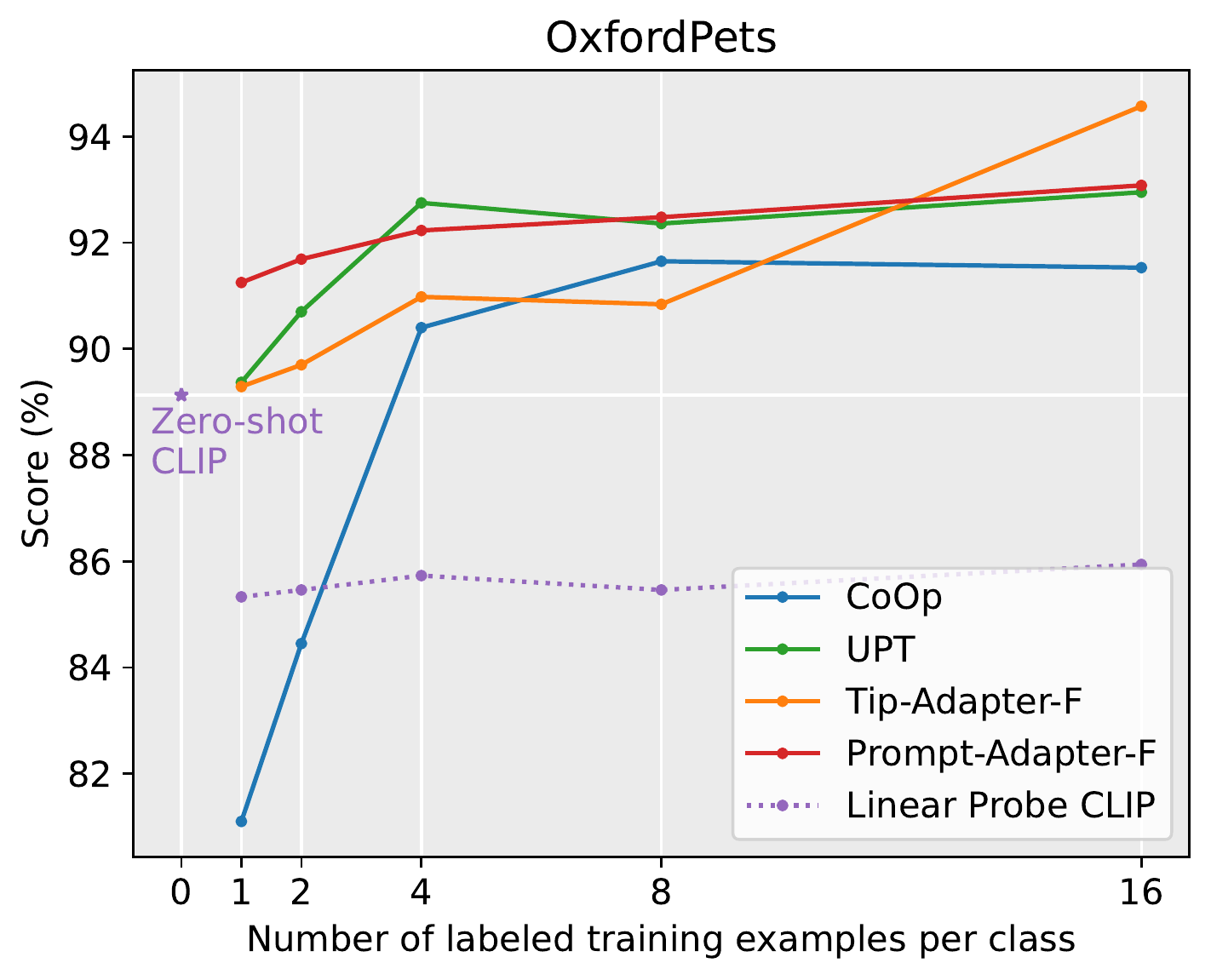}
        \label{fig:b}
    \end{minipage}
}
\caption{Few-shot image classification of 11 datasets.}
\label{fig:fewshot}
\end{figure*}

\begin{table*}[htb]
\centering
\label{tabel-initialize}
\caption{Ablation study of different initialization ways for prompt learning.}
\resizebox{\textwidth}{10mm}{
\begin{tabular}{lllllllllllll}
\hline
                         & Oxford\_Pets & Flowers102 & FGVCAircraft & DTD   & EuroSAT & StanfordCars & Food101 & SUN397 & Caltech101 & UCF101 & ImageNet & Average \\ \hline
Handcrafted   prompt     & 89.21        & 71.34      & 24.72        & 44.39 & 47.60   & 65.32        & 86.06   & 62.50  & 92.94      & 66.75  & 66.73    & 65.23   \\
Random   initialization  & 92.53        & 96.47      & 42.91        & 68.50 & 80.87   & 83.09        & 87.21   & 75.29  & 95.77      & 82.24  & 71.92    & 79.71   \\
Mannul   initialization  & 91.53        & 96.40      & 40.30        & 69.47 & 84.00   & 79.20        & 85.00   & 74.43  & 95.70      & 82.40  & 71.60    & 79.09   \\
Pretrained   initization & 91.74        & 97.20      & 44.91        & 70.21 & 86.02   & 82.20        & 87.37   & 75.78  & 95.98      & 83.45  & 72.23    & 80.64   \\ \hline
\end{tabular}}
\end{table*}

\begin{table*}[htb]
\centering
\label{table:train}
\caption{Ablation study of training strategies.}
\resizebox{\textwidth}{6mm}{
\begin{tabular}{lllllllllllll}
\hline
                    &                  & Oxford\_Pets & Flowers102 & FGVCAircraft & DTD   & EuroSAT & StanfordCars & Food101 & SUN397 & Caltech101 & UCF101 & Average \\ \hline
Joint Training      & Prompt-Adapter-F & 76.25        & 84.37      & 33.61        & 62.40 & 76.68   & 55.42        & 60.73   & 63.90  & 84.13      & 75.53  & 67.30   \\
Separate   Training & Prompt-Adapter-F & 91.66        & 97.89      & 48.39        & 71.28 & 85.37   & 82.91        & 86.35   & 75.56  & 95.74      & 84.43  & 81.96   \\ \hline
\end{tabular}}
\end{table*}

\textbf{Few-shot Learning Results }
We first report results on the 20-shot image setting for the Prompt-Adapter network. After training, we freeze the prompt and fine-tune the learnable cache features. We also use 20-shot images to train other baseline methods. The evaluation results are presented in Table \ref{table:main}, which shows that our Prompt Adapter achieves an average classification accuracy of 78.44\%, which is 1.1\% higher than the Tip-Adapter's (77.35\%) accuracy. This improvement can be attributed to the pre-trained prompt that contains rich text knowledge. Furthermore, when finetuning the learnable cache features with Prompt-Adapter-F, the classification accuracy reaches 81.52\%, which is 0.06\% higher than Tip-Adapter-F's (81.46\%) and Unified Prompt Learning's (81.44\%) accuracy. Remarkably, our Prompt-Adapter-F outperforms the current state-of-the-art methods even when trained on 20-shot images.

For experiments with 1, 2, 4, and 8, 16 shot settings, we plot out the learning curves for different methods in Figure \ref{fig:fewshot}. By looking at the average results over 11 datasets, we find that our method Prompt-Adapter-F achieves the highest accuracy among all the few-shot settings. Specifically, our method outperforms CoOp \cite{coop}, UPT \cite{upt}, and Tip-Adapter-F \cite{tip-adpter} in the 1-shot, 2 shots, 4 shots, and 8 shots experiments. These results indicate that our method can effectively handle scenarios with extremely limited data availability. Notably, our method has demonstrated superior performance compared to UPT \cite{upt}. Because UPT needs 200 epochs of training time to achieve such performance while our method only needs 20 epochs of training time. This phenomenon further highlighting the efficacy of our approach. These findings support the claim that our Prompt Adapter approach is a promising technique for few-shot image classification tasks.

In addition to achieving high accuracy on the average classification accuracy across 11 datasets, our method also outperforms other state-of-the-art methods on individual datasets. Specifically, on datasets such as Flowers102 \cite{flowers102}, FGVCAircraft \cite{fgv}, DTD \cite{dtd}, StanfordCars \cite{standfordcars}, Food101 \cite{food101}, SUN397 \cite{sun397}, Caltech101 \cite{caltech101}, and UCF101 \cite{ucf101}, the proposed Prompt-Adapter-F demonstrates superior performance compared to other methods by significant margins.  Only on some datasets like EuroSAT \cite{eurosat} and OxfordPets \cite{oxfordpets} our method has some slight decreases in accuracy compared with other methods. We claim the reason may be these two datasets have high intra-class visual feature variance \cite{upt}, making text prompt adaption difficult.

In summary, the experimental results demonstrate that our Prompt-Adapter approach is effective and robust for few-shot image classification tasks. The pre-trained prompt and learnable cache features synergistically contribute to the superior performance of our method.

\subsection{Ablation Study} 
In this study, we investigate the effect of initialization and training strategies on the performance of a neural network designed for language generation tasks. Specifically, we evaluate three different initialization methods and two training strategies for the network. 

Regarding the initialized way of the prompt, we test three methods, including random initialization, manual initialization, and pre-trained prompt initialization. Manual initialization means using the embeddings of “a photo of a” to initialize the context vectors \cite{coop}. Both random initialization and manual initialization are single-task initialization ways. And we select the handcraftd prompt as a baseline to compare with. Our results in Table 3 indicate that pre-trained prompt initialization achieves 80.64\% accuracy, and outperform the other methods on all three settings. This method involves first training the prompt on multiple datasets to learn the joint representation of the prompt and then adapting it to single-task training. The superiority of this method suggests that pre-training the prompt can effectively capture the underlying features of the language and enhance the model's performance.

In terms of the training strategy of the network, we employ two approaches, namely joint training and separate training. Our results in Tabel 4 reveal that separate training achieved 81.96\% accuracy on average, better than joint training on the tested datasets. This finding can be attributed to the fact that joint training tends to balance the learnable prompt and learnable cache features, which may lead to a degradation in prompt learning. On the other hand, separate training can better capture the prompt features and enhance the model's ability to generate high-quality language.

Overall, our study provides insights into the impact of initialization and training strategies on the performance of neural networks for language generation tasks. The results highlight the importance of pre-training the prompt and separate training strategy.

\section{Conclusion}
Based on the results of our study, we conclude that the proposed prompt-based adaptation method is an effective approach for efficiently adapting large vision and language models to downstream tasks. By leveraging the strong feature prior knowledge from the cache model and learned text prompt, our method outperforms state-of-the-art approaches on 11 few shots image classification tasks. We believe that our findings can contribute to the community's ongoing efforts to improve the efficiency and effectiveness of vision and language models for a wide range of downstream tasks.

\newpage
{\small
\bibliographystyle{ieee_fullname}
\bibliography{egbib}

\begin{thebibliography}{10}\itemsep=-1pt

\bibitem{afrasiyabi2022matching}
Arman Afrasiyabi, Hugo Larochelle, Jean-Fran{\c{c}}ois Lalonde, and Christian
  Gagn{\'e}.
\newblock Matching feature sets for few-shot image classification.
\newblock In {\em Proceedings of the IEEE/CVF Conference on Computer Vision and
  Pattern Recognition}, pages 9014--9024, 2022.

\bibitem{visual-prompt}
Hyojin Bahng, Ali Jahanian, Swami Sankaranarayanan, and Phillip Isola.
\newblock Exploring visual prompts for adapting large-scale models.
\newblock {\em arXiv preprint arXiv:2203.17274}, 1(3):4, 2022.

\bibitem{food101}
Lukas Bossard, Matthieu Guillaumin, and Luc Van~Gool.
\newblock Food-101--mining discriminative components with random forests.
\newblock In {\em Computer Vision--ECCV 2014: 13th European Conference, Zurich,
  Switzerland, September 6-12, 2014, Proceedings, Part VI 13}, pages 446--461.
  Springer, 2014.

\bibitem{cai2018memory}
Qi Cai, Yingwei Pan, Ting Yao, Chenggang Yan, and Tao Mei.
\newblock Memory matching networks for one-shot image recognition.
\newblock In {\em Proceedings of the IEEE conference on computer vision and
  pattern recognition}, pages 4080--4088, 2018.

\bibitem{dtd}
Mircea Cimpoi, Subhransu Maji, Iasonas Kokkinos, Sammy Mohamed, and Andrea
  Vedaldi.
\newblock Describing textures in the wild.
\newblock In {\em Proceedings of the IEEE conference on computer vision and
  pattern recognition}, pages 3606--3613, 2014.

\bibitem{imagenet}
Jia Deng, Wei Dong, Richard Socher, Li-Jia Li, Kai Li, and Li Fei-Fei.
\newblock Imagenet: A large-scale hierarchical image database.
\newblock In {\em 2009 IEEE conference on computer vision and pattern
  recognition}, pages 248--255. Ieee, 2009.

\bibitem{deng2020meta}
Shumin Deng, Ningyu Zhang, Jiaojian Kang, Yichi Zhang, Wei Zhang, and Huajun
  Chen.
\newblock Meta-learning with dynamic-memory-based prototypical network for
  few-shot event detection.
\newblock In {\em Proceedings of the 13th International Conference on Web
  Search and Data Mining}, pages 151--159, 2020.

\bibitem{dhillon2019baseline}
Guneet~S Dhillon, Pratik Chaudhari, Avinash Ravichandran, and Stefano Soatto.
\newblock A baseline for few-shot image classification.
\newblock {\em arXiv preprint arXiv:1909.02729}, 2019.

\bibitem{caltech101}
Li Fei-Fei, Rob Fergus, and Pietro Perona.
\newblock Learning generative visual models from few training examples: An
  incremental bayesian approach tested on 101 object categories.
\newblock In {\em 2004 conference on computer vision and pattern recognition
  workshop}, pages 178--178. IEEE, 2004.

\bibitem{clip-adapter}
Peng Gao, Shijie Geng, Renrui Zhang, Teli Ma, Rongyao Fang, Yongfeng Zhang,
  Hongsheng Li, and Yu Qiao.
\newblock Clip-adapter: Better vision-language models with feature adapters.
\newblock {\em arXiv preprint arXiv:2110.04544}, 2021.

\bibitem{prompt1}
Tianyu Gao, Adam Fisch, and Danqi Chen.
\newblock Making pre-trained language models better few-shot learners.
\newblock {\em arXiv preprint arXiv:2012.15723}, 2020.

\bibitem{downstream-seg}
Golnaz Ghiasi, Xiuye Gu, Yin Cui, and Tsung-Yi Lin.
\newblock Open-vocabulary image segmentation.
\newblock {\em arXiv preprint arXiv:2112.12143}, 2021.

\bibitem{od-gu2021open}
Xiuye Gu, Tsung-Yi Lin, Weicheng Kuo, and Yin Cui.
\newblock Open-vocabulary object detection via vision and language knowledge
  distillation.
\newblock {\em arXiv preprint arXiv:2104.13921}, 2021.

\bibitem{eurosat}
Patrick Helber, Benjamin Bischke, Andreas Dengel, and Damian Borth.
\newblock Eurosat: A novel dataset and deep learning benchmark for land use and
  land cover classification.
\newblock {\em IEEE Journal of Selected Topics in Applied Earth Observations
  and Remote Sensing}, 12(7):2217--2226, 2019.

\bibitem{parameter2}
Neil Houlsby, Andrei Giurgiu, Stanislaw Jastrzebski, Bruna Morrone, Quentin
  De~Laroussilhe, Andrea Gesmundo, Mona Attariyan, and Sylvain Gelly.
\newblock Parameter-efficient transfer learning for nlp.
\newblock In {\em International Conference on Machine Learning}, pages
  2790--2799. PMLR, 2019.

\bibitem{jia2021scaling}
Chao Jia, Yinfei Yang, Ye Xia, Yi-Ting Chen, Zarana Parekh, Hieu Pham, Quoc Le,
  Yun-Hsuan Sung, Zhen Li, and Tom Duerig.
\newblock Scaling up visual and vision-language representation learning with
  noisy text supervision.
\newblock In {\em International Conference on Machine Learning}, pages
  4904--4916. PMLR, 2021.

\bibitem{vpt}
Menglin Jia, Luming Tang, Bor-Chun Chen, Claire Cardie, Serge Belongie, Bharath
  Hariharan, and Ser-Nam Lim.
\newblock Visual prompt tuning.
\newblock In {\em Computer Vision--ECCV 2022: 17th European Conference, Tel
  Aviv, Israel, October 23--27, 2022, Proceedings, Part XXXIII}, pages
  709--727. Springer, 2022.

\bibitem{downstream-video}
Chen Ju, Tengda Han, Kunhao Zheng, Ya Zhang, and Weidi Xie.
\newblock Prompting visual-language models for efficient video understanding.
\newblock In {\em Computer Vision--ECCV 2022: 17th European Conference, Tel
  Aviv, Israel, October 23--27, 2022, Proceedings, Part XXXV}, pages 105--124.
  Springer, 2022.

\bibitem{koch2015siamese}
Gregory Koch, Richard Zemel, Ruslan Salakhutdinov, et~al.
\newblock Siamese neural networks for one-shot image recognition.
\newblock In {\em ICML deep learning workshop}, volume~2. Lille, 2015.

\bibitem{standfordcars}
Jonathan Krause, Michael Stark, Jia Deng, and Li Fei-Fei.
\newblock 3d object representations for fine-grained categorization.
\newblock In {\em Proceedings of the IEEE international conference on computer
  vision workshops}, pages 554--561, 2013.

\bibitem{krizhevsky2017imagenet}
Alex Krizhevsky, Ilya Sutskever, and Geoffrey~E Hinton.
\newblock Imagenet classification with deep convolutional neural networks.
\newblock {\em Communications of the ACM}, 60(6):84--90, 2017.

\bibitem{parameter1}
Brian Lester, Rami Al-Rfou, and Noah Constant.
\newblock The power of scale for parameter-efficient prompt tuning.
\newblock {\em arXiv preprint arXiv:2104.08691}, 2021.

\bibitem{elevater}
Chunyuan Li, Haotian Liu, Liunian~Harold Li, Pengchuan Zhang, Jyoti Aneja,
  Jianwei Yang, Ping Jin, Yong~Jae Lee, Houdong Hu, Zicheng Liu, et~al.
\newblock Elevater: A benchmark and toolkit for evaluating language-augmented
  visual models.
\newblock {\em arXiv preprint arXiv:2204.08790}, 2022.

\bibitem{prompt2}
Xiang~Lisa Li and Percy Liang.
\newblock Prefix-tuning: Optimizing continuous prompts for generation.
\newblock {\em arXiv preprint arXiv:2101.00190}, 2021.

\bibitem{liu2022few}
Ying Liu, Hengchang Zhang, Weidong Zhang, Guojun Lu, Qi Tian, and Nam Ling.
\newblock Few-shot image classification: Current status and research trends.
\newblock {\em Electronics}, 11(11):1752, 2022.

\bibitem{prompt3}
Yuning Lu, Jianzhuang Liu, Yonggang Zhang, Yajing Liu, and Xinmei Tian.
\newblock Prompt distribution learning.
\newblock In {\em Proceedings of the IEEE/CVF Conference on Computer Vision and
  Pattern Recognition}, pages 5206--5215, 2022.

\bibitem{fgv}
Subhransu Maji, Esa Rahtu, Juho Kannala, Matthew Blaschko, and Andrea Vedaldi.
\newblock Fine-grained visual classification of aircraft.
\newblock {\em arXiv preprint arXiv:1306.5151}, 2013.

\bibitem{flowers102}
Maria-Elena Nilsback and Andrew Zisserman.
\newblock Automated flower classification over a large number of classes.
\newblock In {\em 2008 Sixth Indian Conference on Computer Vision, Graphics \&
  Image Processing}, pages 722--729. IEEE, 2008.

\bibitem{oxfordpets}
Omkar~M Parkhi, Andrea Vedaldi, Andrew Zisserman, and CV Jawahar.
\newblock Cats and dogs.
\newblock In {\em 2012 IEEE conference on computer vision and pattern
  recognition}, pages 3498--3505. IEEE, 2012.

\bibitem{clip}
Alec Radford, Jong~Wook Kim, Chris Hallacy, Aditya Ramesh, Gabriel Goh,
  Sandhini Agarwal, Girish Sastry, Amanda Askell, Pamela Mishkin, Jack Clark,
  et~al.
\newblock Learning transferable visual models from natural language
  supervision.
\newblock In {\em International conference on machine learning}, pages
  8748--8763. PMLR, 2021.

\bibitem{ren2018meta}
Mengye Ren, Eleni Triantafillou, Sachin Ravi, Jake Snell, Kevin Swersky,
  Joshua~B Tenenbaum, Hugo Larochelle, and Richard~S Zemel.
\newblock Meta-learning for semi-supervised few-shot classification.
\newblock {\em arXiv preprint arXiv:1803.00676}, 2018.

\bibitem{shen2022multitask}
Sheng Shen, Shijia Yang, Tianjun Zhang, Bohan Zhai, Joseph~E Gonzalez, Kurt
  Keutzer, and Trevor Darrell.
\newblock Multitask vision-language prompt tuning.
\newblock {\em arXiv preprint arXiv:2211.11720}, 2022.

\bibitem{snell2017prototypical}
Jake Snell, Kevin Swersky, and Richard Zemel.
\newblock Prototypical networks for few-shot learning.
\newblock {\em Advances in neural information processing systems}, 30, 2017.

\bibitem{ucf101}
Khurram Soomro, Amir~Roshan Zamir, and Mubarak Shah.
\newblock Ucf101: A dataset of 101 human actions classes from videos in the
  wild.
\newblock {\em arXiv preprint arXiv:1212.0402}, 2012.

\bibitem{sung2018learning}
Flood Sung, Yongxin Yang, Li Zhang, Tao Xiang, Philip~HS Torr, and Timothy~M
  Hospedales.
\newblock Learning to compare: Relation network for few-shot learning.
\newblock In {\em Proceedings of the IEEE conference on computer vision and
  pattern recognition}, pages 1199--1208, 2018.

\bibitem{tian2020rethinking}
Yonglong Tian, Yue Wang, Dilip Krishnan, Joshua~B Tenenbaum, and Phillip Isola.
\newblock Rethinking few-shot image classification: a good embedding is all you
  need?
\newblock In {\em Computer Vision--ECCV 2020: 16th European Conference,
  Glasgow, UK, August 23--28, 2020, Proceedings, Part XIV 16}, pages 266--282.
  Springer, 2020.

\bibitem{sun397}
Jianxiong Xiao, James Hays, Krista~A Ehinger, Aude Oliva, and Antonio Torralba.
\newblock Sun database: Large-scale scene recognition from abbey to zoo.
\newblock In {\em 2010 IEEE computer society conference on computer vision and
  pattern recognition}, pages 3485--3492. IEEE, 2010.

\bibitem{od-zang2022open}
Yuhang Zang, Wei Li, Kaiyang Zhou, Chen Huang, and Chen~Change Loy.
\newblock Open-vocabulary detr with conditional matching.
\newblock In {\em Computer Vision--ECCV 2022: 17th European Conference, Tel
  Aviv, Israel, October 23--27, 2022, Proceedings, Part IX}, pages 106--122.
  Springer, 2022.

\bibitem{upt}
Yuhang Zang, Wei Li, Kaiyang Zhou, Chen Huang, and Chen~Change Loy.
\newblock Unified vision and language prompt learning.
\newblock {\em arXiv preprint arXiv:2210.07225}, 2022.

\bibitem{tip-adpter}
Renrui Zhang, Wei Zhang, Rongyao Fang, Peng Gao, Kunchang Li, Jifeng Dai, Yu
  Qiao, and Hongsheng Li.
\newblock Tip-adapter: Training-free adaption of clip for few-shot
  classification.
\newblock In {\em Computer Vision--ECCV 2022: 17th European Conference, Tel
  Aviv, Israel, October 23--27, 2022, Proceedings, Part XXXV}, pages 493--510.
  Springer, 2022.

\bibitem{coop}
Kaiyang Zhou, Jingkang Yang, Chen~Change Loy, and Ziwei Liu.
\newblock Learning to prompt for vision-language models.
\newblock {\em International Journal of Computer Vision}, 130(9):2337--2348,
  2022.

\bibitem{od-zhou2022detecting}
Xingyi Zhou, Rohit Girdhar, Armand Joulin, Philipp Kr{\"a}henb{\"u}hl, and
  Ishan Misra.
\newblock Detecting twenty-thousand classes using image-level supervision.
\newblock In {\em Computer Vision--ECCV 2022: 17th European Conference, Tel
  Aviv, Israel, October 23--27, 2022, Proceedings, Part IX}, pages 350--368.
  Springer, 2022.

\end{thebibliography}
}

\end{document}